\documentclass[letterpaper]{article} 
\usepackage{aaai25}  
\usepackage{times}  
\usepackage{helvet}  
\usepackage{courier}  
\usepackage[hyphens]{url}  
\usepackage{graphicx} 
\usepackage{natbib}  
\usepackage{caption} 
\usepackage{algorithm}
\usepackage{algorithmic}
\usepackage[table,xcdraw]{xcolor}

\usepackage{newfloat}
\usepackage{listings}
\DeclareCaptionStyle{ruled}{labelfont=normalfont,labelsep=colon,strut=off} 
\floatstyle{ruled}
\newfloat{listing}{tb}{lst}{}
\floatname{listing}{Listing}
\title{Frequency-Aware Density Control via Reparameterization for High-Quality Rendering of 3D Gaussian Splatting}
\author {
    Zhaojie Zeng\textsuperscript{\rm 1},
    Yuesong Wang\textsuperscript{\rm 1}\thanks{Corresponding author.},
    Lili Ju\textsuperscript{\rm 2},
    Tao Guan\textsuperscript{\rm 1}
}
\affiliations {
    \textsuperscript{\rm 1}School of Computer Science \& Technology, Huazhong University of Science and Technology,  China \\
    \textsuperscript{\rm 2}Department of Mathematics, University of South Carolina, USA \\
    zhaojiezeng@hust.edu.cn, yuesongwang@hust.edu.cn, ju@math.sc.edu, qd\_gt@hust.edu.cn
}

\begin{document}

\maketitle

\begin{abstract}
By adaptively controlling the density and generating more Gaussians in regions with high-frequency information, 3D Gaussian Splatting (3DGS) can better represent scene details. From the signal processing perspective, representing details usually needs more Gaussians with relatively smaller scales. However, 3DGS currently lacks an explicit constraint linking the density and scale of 3D Gaussians across the domain, leading to 3DGS using improper-scale Gaussians to express frequency information, resulting in the loss of accuracy. In this paper, we propose to establish a direct relation between density and scale through the reparameterization of the scaling parameters and ensure the consistency between them via explicit constraints (i.e., density responds well to changes in frequency). Furthermore, we develop a frequency-aware density control strategy, consisting of densification and deletion, to improve representation quality with fewer Gaussians. A dynamic threshold encourages densification in high-frequency regions, while a scale-based filter deletes Gaussians with improper scale. Experimental results on various datasets demonstrate that our method outperforms existing state-of-the-art methods quantitatively and qualitatively.
\end{abstract}

%
\begin{links}
    \link{Code}{https://github.com/whoiszzj/FDS-GS}
\end{links}

\begin{figure}[!htb]
  \includegraphics[width=\linewidth]{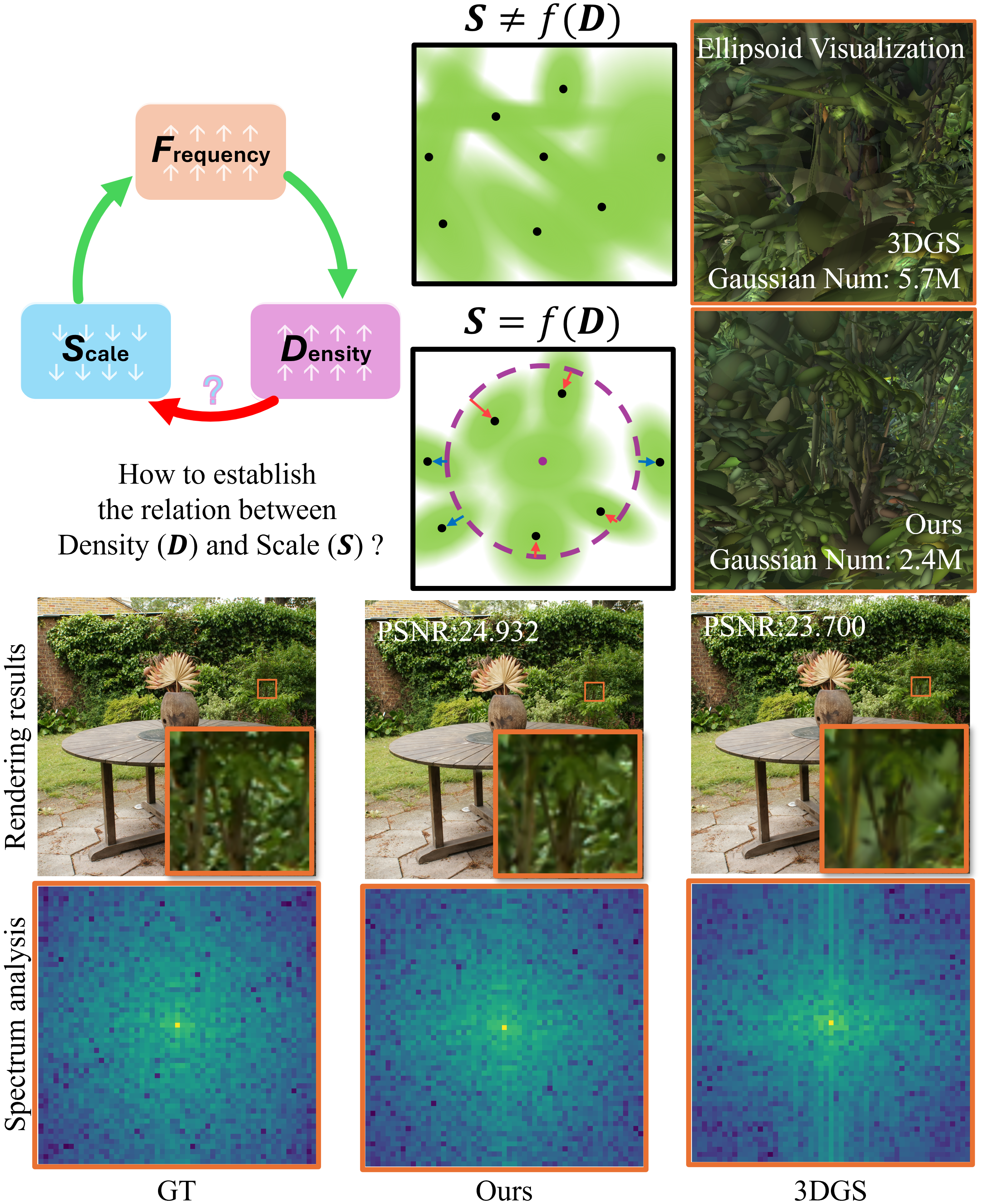}
  \caption{
   By linking the density and scale (the purple circle corresponds to the scale, constrained by the density), our FDS-GS can better reflect the frequency changes than 3DGS via density control (more similar to the GT's spectrum), achieving better render quality with fewer Gaussians. 
  }
  \label{fig:teaser}
\end{figure}
\section{Introduction}
\label{sec:introduction}

Novel View Synthesis (NVS) is a significant task in computer vision, aiming to generate images from arbitrary viewpoints based on scene representations reconstructed from captured images. Neural Field Rendering (NeRF)~\cite{mildenhall2020nerf} has emerged as a leading solution for NVS, renowned for its realistic rendering quality, and has inspired numerous subsequent works~\cite{barron2021mip, barron2022mip360, barron2023zipnerf, kaizhang2020, chen2023mobilenerf}. However, 3D Gaussian Splatting (3DGS)~\cite{kerbl20233d}  has recently gained attention for its combination of high-quality rendering and speed, offering a novel point-based real-time rendering solution for NVS applications in the VR and AR industries~\cite{xu2023vr, deng2022fov, li2018critical}.

In addition to the flexibility offered by organizing a set of 3D Gaussian primitives (termed Gaussians in the following), each defined by position, opacity, spherical harmonic, and covariance matrix decomposed to scale and rotation, another pivotal innovation of 3DGS is its adaptive density control. As practical NVS demands accurate representation of scene details and sparse points are commonly insufficient for scene coverage and detail representation, using dense point cloud as an initial input is necessary for many point-based methods~\cite{kopanas2021point,zhang2022differentiable,franke2024trips}. The adaptive density control liberates point-based methods from the necessity of relying on dense point clouds as input. Consequently, 3DGS can produce high-quality results by leveraging only sparse points generated for free as part of the Structure-from-Motion (SfM) process~\cite{schoenberger2016sfm}.

The underlying principle of density control can be analyzed within the signal processing framework. According to the Fourier transform, complex signals can be represented as a superposition of sine and cosine waves at different frequencies. Sufficient high-frequency components are necessary to capture complex signals, especially abrupt changes, accurately. Similarly, 3DGS uses alpha blending to render scenes by blending Gaussians of different colors to produce the final output. It requires the superposition of many Gaussians to represent complex color variations and sufficient smaller-scale ones to capture high-frequency details.

In practice, 3DGS typically generates more Gaussians through view space gradients and performs split operations on larger Gaussians. However, due to the extensive parameter space in 3DGS, this process does not always result in appropriate scale, sometimes producing overly large Gaussians in regions with details (i.e., local optimal solution), as illustrated in Fig.~\ref{fig:teaser}. Moreover, the decision to split a Gaussian is often made based on its current scale but without understanding the region it represents. This inflexibility will prevent the generation of sufficiently small Gaussians needed to capture fine details as high-frequency components.  From the above discussions and comparison results shown Fig.~\ref{fig:teaser}, we can identify the following inconsistencies in 3DGS. When dealing with high-frequency regions that represent fine details, 3DGS increases the density of Gaussians through densification. Although there is a tendency for the scale to decrease, the lack of explicit and direct regularization constraints leads to that the Gaussians generated through densification do not fully contribute to representing high-frequency information.

To ensure consistency in \textbf{F}requency, \textbf{D}ensity, and \textbf{S}cale in 3DGS, we propose to establish a direct relation between density and scale across the domain by reparameterization. Specifically, we first quantify the density information at each position as the number of Gaussians in a unit volume centered at the position. 
We then reparameterize the scale and decompose it into four variables, including an absolute scalar and a relative scale vector which respond to its anisotropy in three dimensions. Following the principle that higher-density areas will have smaller scales, making them act as high-frequency components, we establish an explicit constraint between the absolute scale scalar and the density. 
We further introduce a frequency-aware density control to enhance the density quality, i.e., using fewer Gaussians to get better rendering quality. 
Following the framework of 3DGS, we use both densification and deletion strategies for density control. First, we propose a dynamic threshold for densification: we sort the gradients of the Gaussians and adjust the threshold based on their distribution. This adaptive approach provides finer control over densification compared to the fixed threshold used in vanilla 3DGS, making the density more responsive to changes in frequency. Subsequently, we design a scale-based confidence filter to effectively remove overfitting Gaussians with improper scales, further ensuring consistency in frequency, density, and scale. We will call our proposed method ``FDS-GS" for brevity.

In summary, our main contributions are as follows:
\begin{itemize}
    \item We quantify the density and reparameterize the scale of the Gaussian to establish a direct relation between density and scale, based on which we use the density to constrain the covariance of the Gaussian to ensure that the density can better respond to changes in frequency.
    \item We develop a frequency-aware density control strategy consisting of densification with a dynamic threshold and a scale confidence filter, optimizing the density to be more expressive of frequency variations to improve the accuracy of the rendering.
    \item Our method, FDS-GS, outperforms existing state-of-the-art methods by achieving higher rendering quality with a reasonable number of Gaussians. 
\end{itemize}

\section{Related Works}
\paragraph{NeRF-based Rendering.} Different from traditional methods that synthesize novel views based on explicit geometry representations such as meshes or points, recently emerged neural rendering methods utilize a set of MLPs to represent the scenes implicitly. Neural Radiance Field (NeRF)~\cite{mildenhall2020nerf} can be seen as a landmark work in this direction. By encoding a continuous volume with a view-dependent appearance using the parameters of a deep fully-connected neural network, NeRF can achieve both photorealistic rendering quality and low storage pressure. Many subsequent works have been proposed to further advance the progress. Mip-NeRF~\cite{barron2021mip} renders the scene using conical frustums instead of rays to reduce aliasing artifacts. Mip-NeRF360~\cite{barron2022mip360} presents an extension of mip-NeRF to handle unbounded scenes better. Martin-Brualla et al.~\cite{martin2021nerf} introduce a series of extensions to NeRF to make it possible to synthesize novel views using only in-the-wild photographs. On the other hand, NeRF-based methods generally suffer from high consumption for training and rendering times due to the deep MLP structure. To tackle this drawback, KiloNeRF~\cite{reiser2021kilonerf} utilizes thousands of tiny MLPs to replace the deep MLP, which can greatly increase the rendering speed. FastNeRF~\cite{garbin2021fastnerf} applies a graphics-inspired factorization to speed up the computation of the pixel values in the rendered image. Deng et al.~\cite{deng2022depth} introduce depth supervision to guide geometry learning, which allows faster training. Instant-NGP~\cite{InstantNGP} replaces the original encoding with multi-resolution hash encoding and only adopts a smaller MLP, consequently leading to a dramatic reduction of time consumption for both training and rendering. Despite the above efforts, it is currently still hard for NeRF-baed methods to achieve both low time consumption and high rendering quality.
 
\paragraph{3DGS-based Rendering.}
In order to ensure high rendering quality while reducing the time overhead, researchers also have adopted different structures~\cite{fridovich2022plenoxels, chen2022tensorf} other than NeRF, but the results were still unsatisfactory enough until the advent of 3D Gaussian Splatting~\cite{kerbl20233d}. 3DGS is developed upon differentiable point-based rendering methods~\cite{zhang2022differentiable, ruckert2022adop} and treat each 3D point as a 3D Gaussian which allows a more flexible representation regime. Then the novel view is rendered using a splatting strategy~\cite{zwicker2001ewa}. Moreover, 3DGS has designed an adaptive density control strategy, which makes it possible to use as input only sparse point clouds that are generated for free when using SfM to obtain camera parameters. Such feature of 3DGS can greatly reduce the computational cost, allowing it to replace NeRF-based methods in almost any scenario~\cite{kocabas2023hugs, chen2023gaussianeditor, zhou2023drivinggaussian,Matsuki:Murai:etal:CVPR2024,keetha2023splatam}. Given the great potential of 3DGS, more studies have been invested to further improve its performance. Chung et al.~\cite{chung2023depth} introduce monocular depth estimation to supervise additionally to improve the performance given sparse-view inputs. Scaffold-GS~\cite{scaffoldgs} divides the space into voxels, which serve as anchors for nearby 3D Gaussians so that these 3D Gaussians have a soft geometric constraint and the probability of generating outliers is reduced. GaussianPro~\cite{cheng2024gaussianpro} explicitly constrains the positions of the newly generated 3D Gaussians by introducing the idea of depth propagation in Multi-View Stereo. While these works mainly focus on improving the rendering quality by constraining the geometry of the 3DGS to reduce the floaters, other researches~\cite{zhang2024fregs, yu2024mip} have noted the importance of frequency constraints for 3DGS, given that nerf-based methods~\cite{xu2023wavenerf} have demonstrated the importance of learning in the frequency space. Mip-Splatting~\cite{yu2024mip} goes for alias-free 3DGS with frequency constraints, which mainly addresses the problem of inconsistent rendering at different resolutions but does not pay attention to improving the rendering at the original resolution. 

FreGS~\cite{zhang2024fregs} attempted to interpret and optimize 3DGS from a frequency perspective by transferring the image into the frequency domain via a Fourier transform and designing a frequency loss function that encourages the rendered image to be consistent with the training image in the frequency domain. This approach can partially solve the problem of fitting high-frequency information with large-scale Gaussians. Due to the large frequency loss in these regions, the density control will make the large-scale Gaussians further split into smaller ones, thus jumping out of the pit of local optima. On the other hand, further densification in this approach will simultaneously lead to a huge rise in Gaussian numbers, causing giant rendering and storage pressure. In addition, the same spatial region will result in large differences in image-space frequency if it is seen from different viewpoints, i.e., the image-space frequency does not have multi-view consistency. For instance, a region with mid-frequency information might appear as high-frequency from certain viewpoints, resulting in an excessive number of small-scale Gaussians, further causing redundancy.
Therefore, improving the rendering quality of 3DGS from the frequency perspective still remains to be delved into.

\section{Proposed Method}
\subsection{Overview}

\begin{figure*}[ht!]
  \centering
  \includegraphics[width=\linewidth]{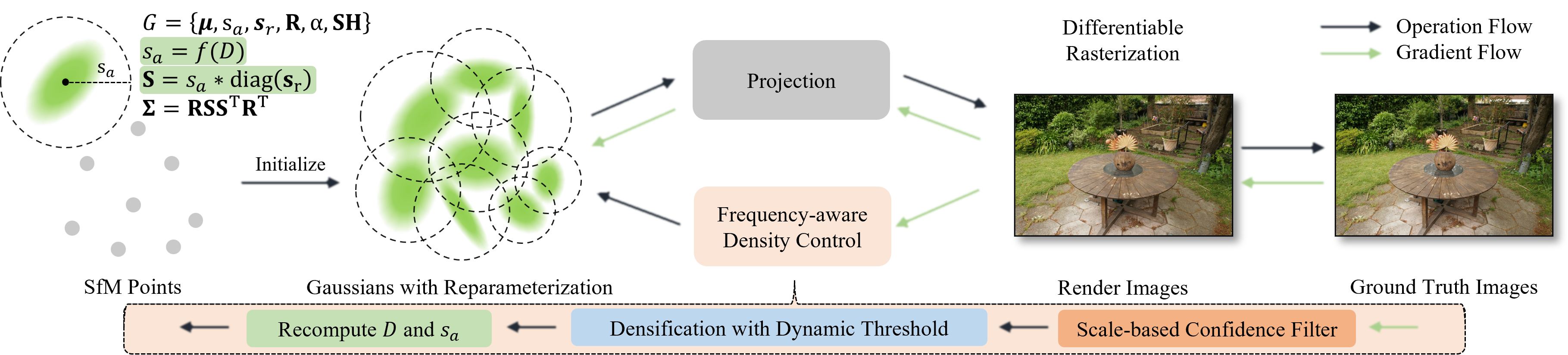}
  \caption{
     The pipeline of the proposed FDS-GS. FDS-GS is an improvement upon the 3DGS. By reparameterizing the scale $\mathbf{S}$ of the Gaussian and quantifying the density as $D$, we establish a direct relation between density and scale and constrain the scale according to the density to better express the frequency changes of the scene. To better control the density to achieve better rendering with fewer Gaussians, every few iterations, we compute the confidence of each Gaussian, delete the Gaussians with low confidence, and then compute a dynamic threshold for densification based on the gradient of the remaining Gaussians. Once the Gaussian distribution has changed, we will then recompute the values of $D$ and $s_a$.
     }
  \label{fig:pipeline}
\end{figure*}

In the following section, we will first briefly review the vanilla 3DGS method. Then, we will delve into the motivations behind building a direct relation between density and scale and explain how it is implemented via reparameterization. After linking the density and scale, we introduce a frequency-aware density control to better capture frequency variations with density changes. This control strategy involves densification with a dynamic threshold to enable the method to respond more swiftly to high-frequency information and a scale-based confidence filter that computes confidence scores for Gaussians and filters them accordingly. The frequency-aware density control process allows us to significantly reduce the number of Gaussians while still achieving high reconstruction quality. Fig.~\ref{fig:pipeline} illustrates the pipeline.

\subsection{Preliminary}
\label{sec:preliminary}

3DGS~\cite{kerbl20233d} represents scenes through a series of anisotropic Gaussian primitives and obtains the rendered images for the corresponding viewpoint through the splatting algorithm. Each primitive is parameterized by a position $ \mathbf{\mu} \in \mathcal{R}^3 $, an opacity $\alpha \in [0, 1]$, a covariance matrix $\mathbf{\Sigma} \in \mathcal{R} ^ {3 \times 3}$ and spherical harmonic ($\mathbf{SH}$) coefficients. The 
probability distribution for each Gaussian $G$ is defined as:
\begin{equation}
    \label{eq:gaussian_distribution}
    G(\mathbf{x}) = e^{-\frac{1}{2} (\mathbf{x} - \mathbf{\mu})^T \mathbf{\Sigma}^{-1} (\mathbf{x} - \mathbf{\mu})},
\end{equation}
where $\mathbf{x} \in \mathcal{R}^3$ is an arbitrary position. To maintain the semi-positive definiteness of the covariance matrix $\mathbf{\Sigma} $, it is often to decompose it into a scaling matrix $\mathbf{S}$ and a rotation matrix $\mathbf{R}$ such that:
\begin{equation}
    \label{eq:decomposition}
    \mathbf{\Sigma} = \mathbf{R} \mathbf{S} \mathbf{S}^T \mathbf{R}^T,
\end{equation}
where $\mathbf{S}$ is a diagonal matrix with positive entries, which can be parameterized by a vector $\mathbf{s} \in \mathcal{(R^+)}^3$, and $\mathbf{R}$ is a rotation matrix, which can be parameterized by a quaternion $\mathbf{q} \in \mathcal{R}^4$.
To render an image for a given viewpoint, the 3D Gaussians $G$ are first projected to get a 2D Gaussians $G'$ according to~\cite{zwicker2001ewa}, and the spherical harmonics are converted to view-dependent color $\mathbf{c}$. Then, the image is rendered by alpha-blending according to the Gaussians' depth order based on the tile. For each pixel with position $\mathbf{x'}$ on the image, the color is acquired by:
\begin{equation}
    \label{eq:alpha_blending}
    \mathbf{C}(\mathbf{x'}) = \sum_{k=1}^{K}{\mathbf{c}_k \alpha_k G'_k(\mathbf{x'}) T_k},
\end{equation}
with
\begin{equation}
     T_k = \prod_{j=1}^{k-1}{(1-\alpha_j G'_j(\mathbf{x'}))},
\end{equation}
where $K$ denotes the number of Gaussians used to render the pixel. Since 3DGS takes sparse points from SfM as input, usually inadequate for scene details representation, a densification strategy is then introduced to create new Gaussians. A key criterion for densification involves accumulating view space gradients. If the gradient exceeds a preset parameter $\tau_{pos}$, the Gaussian will be cloned or split based on the scale of the Gaussian. Cloning duplicates Gaussians while splitting divides large Gaussians into smaller intersecting ones. Newly generated Gaussians inherit most attributes from the original Gaussian seamlessly. Additionally, to control the number of Gaussians, 3DGS suggests resetting opacity every 3000 iterations and periodically removing Gaussians with opacity below a preset parameter $\epsilon_{\alpha}$ or with excessively large scales.

\subsection{Linking the Density and Scale}
\subsubsection{\em Density Estimation and Scale Reparametrization}

As mentioned before, when rendering high-frequency details, 3DGS often increases density in these areas, allowing a greater number of Gaussians to capture the details and reduce discrepancies between the ground truth and rendered images. However, without explicit constraint on the scale of each Gaussian, the extensive solution space associated with the 3DGS parameterization can result in an inappropriate distribution of improper-scale Gaussians, thereby hindering accurate representation of the scene details as demonstrated in Fig.~\ref{fig:teaser}. To address this problem, we introduce an explicit constraint by defining a relation between density and scale.

Our first step is to quantify density information based on the current Gaussian distribution. Here, we let the density refer to the number of points per unit volume at each position. Specifically, for a set of Gaussians $\{G\}$, the density $D$ at a position $\mathbf{x}$ is estimated as follows:
\begin{equation}
    D(\mathbf{x}) = \frac{K(\mathbf{x}, \Delta V,{G})}{\Delta V},
\end{equation}
where $K$ denotes the number of Gaussians in $\{G\}$ which intersect the region described by a unit volume $\Delta V$. The unit volume $\Delta V$ is a small region in 3D space,  centered at a position $\mathbf{x}$. The size of $\Delta V$ is determined by a scale parameter, which could represent the radius of a sphere.

Next, we try to establish a direct relation between density and scale. By reparameterizing the scale vector $\mathbf{s}$ into an absolute scale scalar $s_a > 0$ and a relative scale vector $\mathbf{s}_r \in (0,1)^3$, all of which are learnable, we impose a constraint -- the absolute scale scalar $s_a$ is defined as a function of density, expressed as:
\begin{equation}
    s_a = f(D).
\end{equation}
Then the scaling matrix in Eq.~(\ref{eq:decomposition}) is rewritten as:
\begin{equation}
    \mathbf{S} = {\rm diag}(\mathbf{s}), \quad
    \mathbf{s} = s_a(\mathbf{s}_{r_1}, \mathbf{s}_{r_2}, \mathbf{s}_{r_3}).
\end{equation}

\subsubsection{\em Relation Model and Scale Constraint}
To apply the above definitions to the training optimization process, we first need to compute the density at the center $\mu$ of each Gaussian and then compute the absolute scale scalar $s_a$ corresponding to this density by defining a relation function $f$. Instead of defining a fixed search range for density estimation, which is difficult to be generalizable for all scenarios and computationally expensive, we employ a CUDA-based KDTree~\cite{grandits_geasi_2021} to find the $K$ nearest neighbors of each Gaussian. The weighted distance to these neighbors is used to estimate the local density $D$ for each Gaussian as follows:

\begin{equation}
\label{eq:density}
D (\mu)=\frac{K}{\frac{4}{3} \pi \widetilde{R}^3},
\end{equation}
with
\begin{equation}
    \widetilde{R} = \frac{\sum_{k=1}^{K}{\omega_k d_{\mathbf{q}_k \mathbf{\mu}} }}{\sum_{k=1}^{K}{\omega_k}},
\end{equation}
where $K$ denotes a preset fixed number of neighbor Gaussians for density calculation, $d_{\mathbf{q}_k \mathbf{\mu}}$ is the distance of the $k$-th neighbor Gaussian to the current Gaussian's center $\mathbf{\mu}$, $\omega_k>0$ is a weighting parameter defined by: 
\begin{equation}
    \omega_k = \exp\left(-\left( \frac{d_{\mathbf{q}_k \mathbf{\mu}}  -d_{\mathbf{q}_1 \mathbf{\mu}}}{ median_{\mathbf{\mu}}\{d_{\mathbf{q}_1 \mathbf{\mu}}\}}\right)^2\right),
\end{equation}
where $d_{\mathbf{q}_1 \mathbf{\mu}}$ is the nearest neighbor distance, which represents the distance between the center point of $G$'s nearest neighbor and $\mathbf{\mu}$, and $median_{\mathbf{\mu}}\{d_{\mathbf{q}_1 \mathbf{\mu}}\}$ denotes the median of the nearest neighbor distances for all Gaussians, which corresponds to a scale scalar factor for this scene. With the use of $\omega_k$, $\widetilde{R}$ can be adaptively adjusted according to the scale of the scene and the Gaussian distribution to ensure generalization.

Having quantified the density, we need to calculate the corresponding absolute scale scalar. Intuitively, higher-density regions should have smaller-scale Gaussians and vice versa, indicating a negative correlation. From Eq.~(\ref{eq:density}), we observe a negative correlation between $\widetilde{R}$ and density. Thus, it infers that the $\widetilde{R}$ and the absolute scale scalar $s_a$ are positively correlated. Furthermore, $\widetilde{R}$ and $s_a$ should belong to the same domain in terms of their physical meaning. Therefore, here we simply define the mapping between $D$ and $s_a$ as follows:
\begin{equation}
    s_a = f(D(\mathbf{\mu})) := \widetilde\theta D(\mathbf{\mu}) ^ {-\frac{1}{3}} = {\theta} \widetilde{R},
\end{equation}
where $\theta=\widetilde\theta(\frac{4\pi}{3K})^{1/3} >0$ is a predfined constant parameter related to $K$. To optimize computational efficiency, we recompute the density $D$ only when the Gaussians undergo structural changes, i.e., adding or removing Gaussians. We then compute the corresponding $s_a$ based on $f$ to reset the scale of each Gaussian to impose the explicit constraints. Note that $\mathbf{s}_r$ will not be reset, preserving the original shape of the Gaussian.

\subsection{Frequency-aware Density Control}
By linking the density and scale and imposing explicit constraints on the scale, we now expect to better express the changes in scene frequency by controlling the density, achieving better rendering quality. However, the density control of vanilla 3DGS is not designed from the frequency perspective, so the final density distribution isn't the optimal one to express frequency changes. To resolve this issue, i.e., to use fewer Gaussians to get better rendering quality as much as possible, we design a frequency-aware density control approach. As with 3DGS, our frequency-aware density control also consists of densification and deletion.

\subsubsection{\em Densification with Dynamic Threshold}
The complexity of scenes varies, and achieving the best rendering effect often requires meticulous attention to detail.
While large-scale Gaussians struggle to represent regions with high-frequency information, a sufficient number of small-scale Gaussians can readily convey both low-frequency and high-frequency information.
Given a fixed threshold for densification, it is safest to set it low enough to encourage over-densification of the Gaussians, allowing numerous small-scale Gaussians to fit arbitrary scenarios. However, this naive strategy would cause the explosion of Gaussians and thus is impractical considering resource limitations.
Therefore, we need a wiser threshold that not only ensures enough Gaussian to express the scene details but also avoids redundancy from over-densification.
We propose to use a dynamic gradient threshold approach to resolve the above issue. In each densification process, we calculate the minimum gradient value $grad_{min}$ and the mean gradient value $grad_{mean}$ and then perform histogram statistics between $grad_{min}$ and $3\cdot grad_{mean}$. From largest to smallest, we compute the gradients $grad_{25\%}$ of the top 25\% Gaussian points, compare them with the preset threshold $grad_{preset}$, and obtain the final densification gradient threshold $\tau_{pos}=max(grad_{25\%}, grad_{preset})$.

In the early stage of optimization, we mainly rely on $grad_{25\%}$ to determine the threshold of densification. This scene-dependent threshold encourages higher density in regions currently with higher rendering discrepancy while simultaneously ensuring a controlled and gradual increase in the number of Gaussians. This way prevents an unmanageable explosion in Gaussians, which could impose significant computational burdens. On the other hand, it also could avoid optimization stagnation due to a mismatch between the threshold setting and the scene. After the optimization has stabilized, $grad_{preset}$ is gradually considered more to prevent excessive densification.

\subsubsection{\em Scale-based Confidence Filter}
Although our method can link scale and density to better represent frequency changes, due to the stochastic nature of the splitting operation in the densification and the complexity of the optimization, the Gaussians may not always be closely distributed on the real scene surface. This could result in the density possibly not matching well with the scene, negatively affecting our scale constraints in some regions. Thus, we need to correct the density of these regions. Inspired by the PatchMatch~\cite{bleyer2011patchmatch} method, we design below a scale-based confidence filter, which calculates the photometric consistency of the Gaussian on different views according to the scale, uses it as a basis for determining the Gaussian's confidence and filters out the ones with low confidence.

\begin{figure}[tb]
  \centering
  \includegraphics[width=1.0\linewidth]{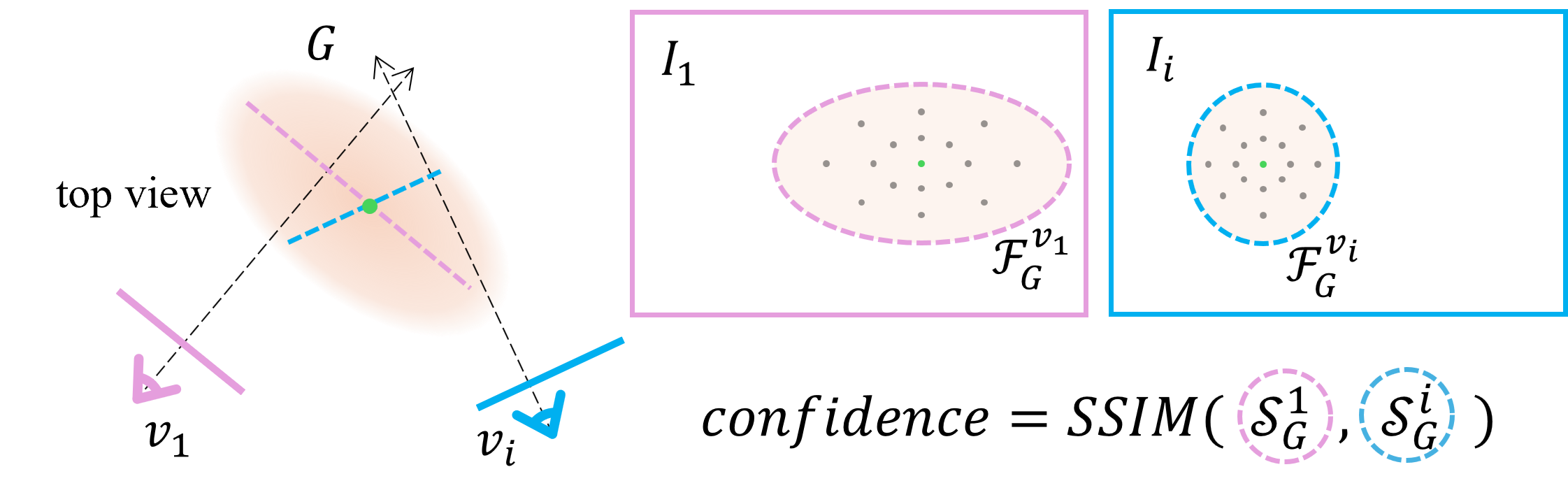}
  \caption{ Confidence Calculation Example. For a given Gaussian $G$, project it onto observable images to obtain the corresponding footprints. We sample several points within the footprints to calculate the SSIM as the confidence.
     }
  \label{fig:confidencefilter}
\end{figure}

\begin{table*}[!htb]
\centering
\resizebox{0.85\linewidth}{!}{
\setlength{\tabcolsep}{6pt}
\begin{tabular}{c|ccc|ccc|ccc}
\hline
Dataset & \multicolumn{3}{c|}{MipNeRF-360} & \multicolumn{3}{c|}{Tanks\&Temples} & \multicolumn{3}{c}{Deep Blending} \\ \hline
Method | Metrics & SSIM$\uparrow$ & PSNR$\uparrow$ & LPIPS$\downarrow$ & SSIM$\uparrow$ & PSNR$\uparrow$ & LPIPS$\downarrow$ & SSIM$\uparrow$ & PSNR$\uparrow$ & LPIPS$\downarrow$ \\ \hline
Instant-NGP & 0.699 & 25.59 & 0.331 & 0.745 & 21.92 & 0.305 & 0.817 & 24.96 & 0.390 \\
Mip-NeRF360 & 0.792 & 27.69 & 0.237 & 0.759 & 22.22 & 0.257 & 0.901 & 29.40 & 0.245 \\ \hline
3DGS & 0.815 & 27.21 & 0.214 & 0.841 & 23.14 & 0.183 & 0.903 & 29.41 & 0.243 \\
3DGS\dag & 0.813 & 27.48 & 0.221 & 0.848 & 23.89 & 0.181 & 0.900 & 29.48 & 0.247 \\
Scaffold-GS\dag & 0.806 & 27.57 & 0.237 & 0.852 & \cellcolor[HTML]{FFCC99}24.51 & 0.189 & \cellcolor[HTML]{FF9999}0.907 & \cellcolor[HTML]{FF9999}29.98 & 0.257 \\
MipSplatting\dag & \cellcolor[HTML]{FFCC99}0.827 & 27.73 & \cellcolor[HTML]{FFCC99}0.190 & \cellcolor[HTML]{FFCC99}0.859 & 23.93 & \cellcolor[HTML]{FFCC99}0.161 & 0.903 & 29.34 & \cellcolor[HTML]{FFCC99}0.239 \\
3D-HGS\dag & 0.813 & \cellcolor[HTML]{FFF6B2}27.84 & 0.218 & \cellcolor[HTML]{FFCC99}0.859 & \cellcolor[HTML]{FF9999}24.65 & \cellcolor[HTML]{FFF6B2}0.173 & \cellcolor[HTML]{FFF6B2}0.904 & 29.44 & 0.244 \\
FreGS & \cellcolor[HTML]{FFF6B2}0.826 & \cellcolor[HTML]{FFCC99}27.85 & \cellcolor[HTML]{FFF6B2}0.209 & 0.849 & 23.96 & 0.178 & \cellcolor[HTML]{FFF6B2}0.904 & \cellcolor[HTML]{FFF6B2}29.93 & \cellcolor[HTML]{FFF6B2}0.240 \\ \hline
Ours & \cellcolor[HTML]{FF9999}0.835 & \cellcolor[HTML]{FF9999}27.95 & \cellcolor[HTML]{FF9999}0.180 & \cellcolor[HTML]{FF9999}0.863 & \cellcolor[HTML]{FFF6B2}24.27 & \cellcolor[HTML]{FF9999}0.141 & \cellcolor[HTML]{FFCC99}0.906 & \cellcolor[HTML]{FF9999}29.98 & \cellcolor[HTML]{FF9999}0.231 \\ \hline
\end{tabular}
}
\caption{ Quantitative comparison of our FDS-GS method with other state-of-the-art methods. The results reproduced by ourselves using the official code are marked with \dag. The best, second best, state-of-the-art best are marked in red, orange, and yellow, respectively.}
\label{tab:mipnerf360}
\end{table*}

Firstly, we need to select several observable views for each Gaussian. For a given view $v_i$, we obtain the contribution $c$ of the Gaussian $G$ under this view by:
\begin{equation}
    c(G, v_i) = \sum_{\mathbf{x}' \in \mathcal{F}_G^{v_i}}{\alpha G'(\mathbf{x}')},
\end{equation}
where $\mathcal{F}_G^{v_i}$ is the footprint of $G$ at view $v_i$ and $G'$ denotes its 2D Gaussian distribution. Then, we employ a CUDA-based Heap Sort to retrieve the top $M>1$ views with the largest contributions for the Gaussian $G$ as the observable views $V^{(M)}$:
\begin{equation}
    V^{(M)} = HeapSort_M({c(G, v_i)\,|i=1,2,...,N_v}),
\end{equation}
where $N_v$ is the total number of views. According to the 2D Gaussian distribution obtained from each projection, we sample 49 points within the footprint and compute their weight, denoted as $\mathcal{S}_G^{i}$ and $\Omega_G^{i}$, as shown in Fig.~\ref{fig:confidencefilter}. Please refer to the supplementary material for more details.

After obtaining the sampling points for the Gaussian $G$ across the observable views $V^{(M)}$, we calculate the SSIM~\cite{wang2004image} between these sampling points:
\begin{equation}
    confidence = \frac{1}{M-1} \sum_{i=2}^{M}{SSIM(\mathcal{S}_G^{1}, \Omega_G^{1},\mathcal{S}_G^{i}, \Omega_G^{i})}.
\end{equation}
where $\mathcal{S}_G^{1} $ and $\Omega_G^{1}$  are from the view with the largest contribution. Please refer to the supplementary material for more details. After obtaining the confidence for each Gaussian (ranging from [0, 1]), we filter out those with extremely poor confidence using a fixed threshold $\tau_c$. 

\section{Experimental Results}

\begin{figure*}[!htb]
  \centering
  \includegraphics[width=\linewidth]{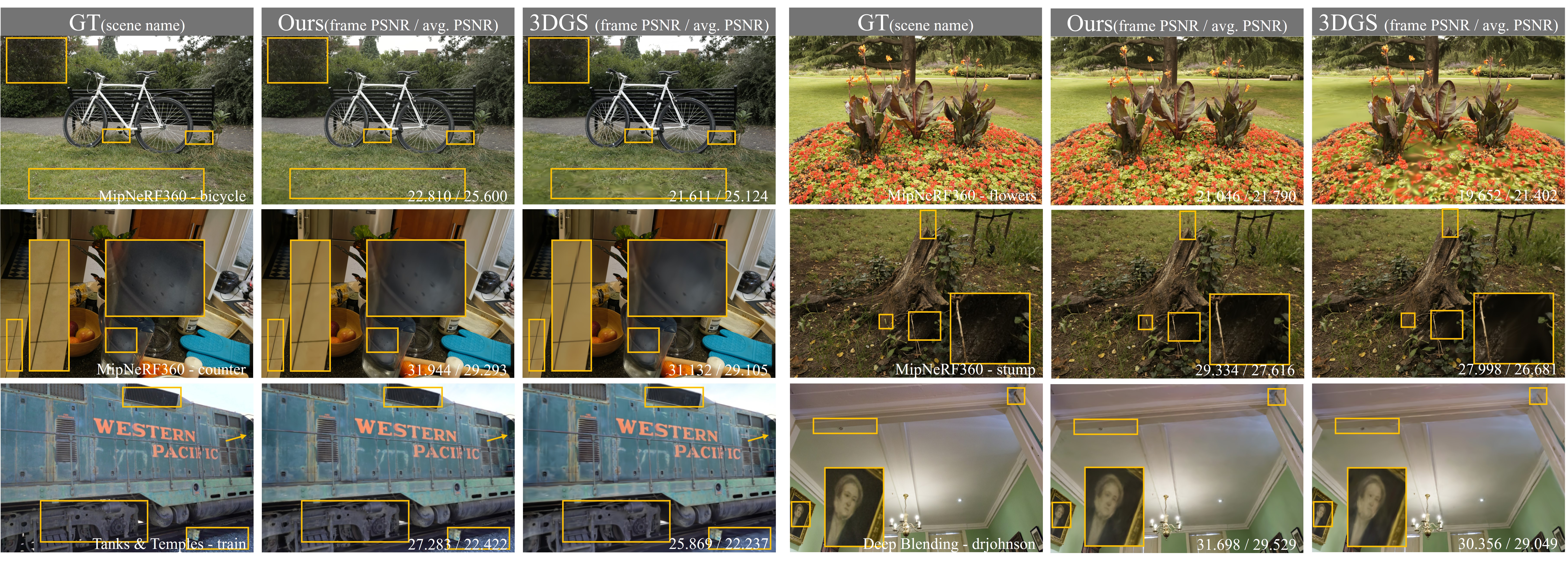}
  \caption{
     Qualitative comparisons of our FDS-GS method with vanilla 3DGS. Our method provides a superior restoration of scene details. GT means the ground truth. Zoom in for more details.
     }
  \label{fig:experiment_vis}
\end{figure*}

\subsection{Implementation}
We select 3DGS~\cite{kerbl20233d} as the baseline and implement our FDS-GS method based on its open-source code. Following 3DGS, we utilize the SfM point cloud obtained from Colmap~\cite{schoenberger2016sfm} for initialization. Except for the first 1000 iterations, densification is executed every 500 iterations, followed by recomputing density and $s_a$.The scale confidence filter is executed every 1000 iterations. The frequency-aware density control persists until reaching 15000 iterations. In all our experiments, we set $ K = 50, \theta = 1.2$ for linking the density and scale, $grad_{preset} = 0.0005$ for dynamic threshold, and $M=2$, $\tau_c = 0.2$ for scale confidence filter.

\subsection{Datasets}
Following 3DGS, we utilize the full set of scenes presented in \textit{Mip-NeRF360}~\cite{barron2022mip360}, two scenes from the \textit{Tanks\&Temples}~\cite{Knapitsch2017} and two scenes provided by the \textit{Deep Blending}~\cite{hedman2018deep} to test our method. We use common metrics (PSNR, SSIM~\cite{wang2004image}, LPIPS~\cite{zhang2018unreasonable}) to quantify the performance. All experiments were performed on an A100 GPU.

\subsection{Quantitative and Qualitative Comparison}
We compare our method with other state-of-the-art methods, including Instant-NGP~\cite{InstantNGP}, MipNeRF360~\cite{barron2022mip360}, 3DGS~\cite{kerbl20233d}, Scaffold-GS~\cite{scaffoldgs}, MipSplatting~\cite{yu2024mip}, 3D-HGS~\cite{li20243d} and FreGS~\cite{zhang2024fregs} on the \textit{MipNeRF360}, \textit{Tanks\&Temples}, and \textit{Deep Blending} datasets. Since some of them do not provide experimental results for the full set of \textit{Mip-NeRF360} in their papers, we have reproduced those on the datasets using the official code provided by the authors with the same coding environment. Quantitative results are shown in Tab.~\ref{tab:mipnerf360}. Our method outperforms other methods on all datasets, further proving its robustness. Some qualitative results are depicted in Fig.~\ref{fig:experiment_vis}. The scenes shown in Fig.~\ref{fig:experiment_vis} are all complex with rich details, which highly tests the method's ability to represent scene details. Due to the lack of explicit scale constraints in 3DGS, it is often difficult to cope with complex frequency changes by controlling the density alone, and thus, the details of the scene are difficult to preserve.
On the other hand, benefiting from linking density and scale, our FDS-GS method can explicitly encourage consistency between them, which reduces the optimization difficulty and leads to better performance in detail restoration. While FreGS also aims to preserve these scene details, as shown in Tab.~\ref{tab:mipnerf360}, its overall performance is not as good as ours. In addition, it usually introduces a large number of redundant Gaussians (as mentioned in its paper, the number of Gaussians on the \textit{stump} set shoots up to 8.1 million, compared with the 4.6 million of the vanilla Gaussian). At the same time, our method has reasonable control over density with the Gaussian number 1 million less than the vanilla 3DGS (3.2 million on \textit{stump}). More specific data will be given in the supplementary material.

\subsection{Ablation Study}
We conduct ablation studies on the \textit{MipNeRF360} full set to demonstrate the effectiveness of the modules we propose.

\subsubsection{\em Linking the Density and Scale}
\begin{figure}[!htb]
  \centering
  \includegraphics[width=1\linewidth]{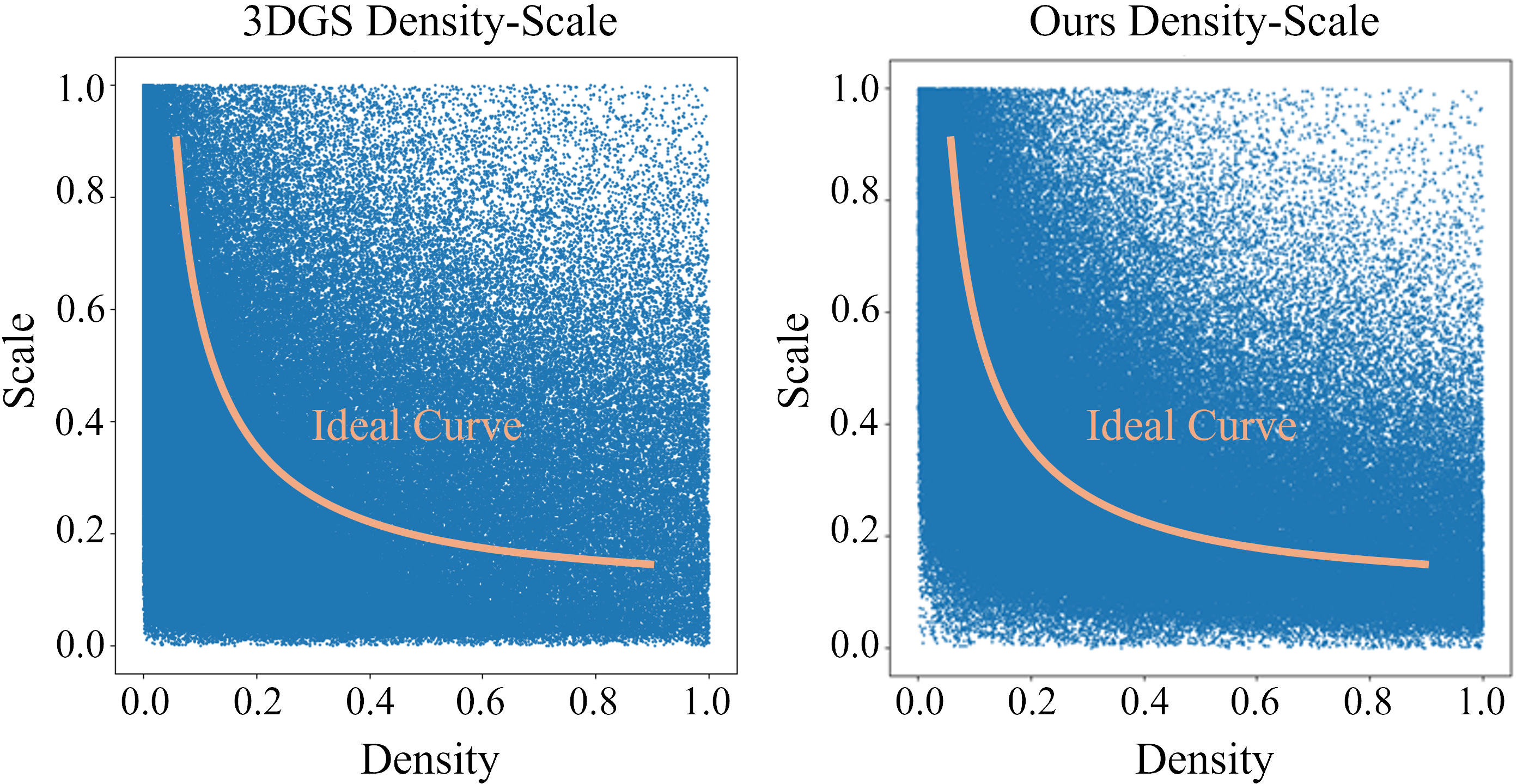}
  \caption{
        The relation between density and scale. We randomly sample 25,000 Gaussians from the point cloud at iteration 30,000 (\textit{MipNeRF360—bicycle}). The \textit{Density} is computed according to Eq.~(\ref{eq:density}), and the \textit{Scale} corresponds to the volume of the Gaussian calculated using the scaling matrix. The ideal curve is obtained by fitting the result. The Gaussians of our methods are distributed around the ideal curve, while those of 3DGS show a more diffuse distribution.
     }
  \label{fig:scale_density_relation}
\end{figure}

In this ablation experiment, we eliminate the scale constraint provided by linking density and scale to prove its effectiveness. From Tab.~\ref{tab:ablation}, we can observe that the number of Gaussians is slightly increased compared to the full-version method, yet the rendering quality is significantly degraded. This is a good indication that solely controlling the density can not maintain a good representation of the frequency changes in the scene. With our scale constraint, we can obtain a better rendering result by simultaneously controlling both density and scale. As illustrated in Fig.~\ref{fig:scale_density_relation}, we visualized the relation between density and scale through a scatter plot. It can be observed that the \textit{Density} and \textit{Scale} are in negative correlation. Our FDS-GS method and 3DGS both have this trend, but our method displays a tighter distribution, indicating the effectiveness of the constraint between scale and density.

\subsubsection{\em Densification with Dynamic Threshold}
We now test the effectiveness of the dynamic threshold in densification. We compare it with the fixed gradient threshold, $\tau_{pos} = 0.0005$. As seen from Tab.~\ref{tab:ablation}, the rendering quality achieved with a fixed threshold is nearly the same as with our dynamic threshold, aside from minor fluctuations. However, the number of Gaussians increases by almost 1 million, indicating significant over-densification. In other words, Gaussians with excessive density are used to represent regions with minimal frequency changes. This redundant representation, caused by the mismatch between density and frequency, leads to unnecessary rendering and storage overhead.

\subsubsection{\em Scale-based Confidence Filter}
In this experiment, we replace the scale-based confidence filter with the original filter used in the vanilla 3DGS. As shown in Tab.~\ref{tab:ablation}, using the original filter causes the number of Gaussians to skyrocket, which not only fails to improve rendering quality but also degrades it since incorrect density misguides the scale. Our proposed filter more effectively controls density and, when combined with scale, leads to better rendering results.

\begin{table}[!htb]
\centering
\resizebox{1.0\linewidth}{!}{
\setlength{\tabcolsep}{2pt}
\begin{tabular}{l|cccc}
\hline
\multicolumn{1}{c|}{Method -- Metrics} & SSIM$\uparrow$ & PSNR$\uparrow$ & LPIPS$\downarrow$ & Gauss.Num.$\downarrow$ \\ \hline
3DGS                                   & 0.813          & 27.48          & 0.221             & 3174k                  \\ \hline
+ D\&S linking                         & 0.827          & 27.65          & 0.178             & 5442k                  \\
+ D\&S linking + dyna. thres.        & 0.831          & 27.75          & 0.178             & 3770k                  \\
+ D\&S linking + conf. filter          & 0.836          & 27.97          & 0.175             & 3123k                  \\
Full                                   & 0.835          & 27.95          & 0.180             & 2370k                  \\ \hline
\end{tabular}
}
\caption{ Ablation study. We sequentially add scale-constraint (+ D\&S linking), dynamic threshold for densification (+ dyna. thres.), and scale-based confidence filter (+ conf. filter) to demonstrate their effectiveness.
}
\label{tab:ablation}

\end{table}

\section{Conclusion}
In this paper, we develop FDS-GS, which can significantly enhance the performance of 3DGS by linking the density and scale to better capture frequency changes. We reparameterize scale, quantify density, and establish a direct relation between them, using density to constrain scale for consistency. This approach generates more Gaussians with small scales when high-frequency information is required, improving rendering quality. Additionally, we design a frequency-aware density control mechanism to optimize density for accurate scene representation. Experimental results on various datasets show that FDS-GS performs better than existing 3DGS-based approaches.

\section{Acknowledgements}
 This work is supported by the Natural Science Foundation of China under Grant 62302174. The computation is completed in the HPC Platform of Huazhong University of Science and Technology. We also thank Farsee2 Technology Ltd for providing devices to support the validation of our method.

\bibliography{aaai25}

\end{document}


\maketitle

\section{Validation for relation between Frequency, Density, and Scale}
In the paper's introduction, we've mentioned that "It requires the superposition of many Gaussians to represent complex color variations and sufficient smaller-scale ones to capture high-frequency details." This implies that to recover high-frequency information, it is necessary to have a sufficient number of Gaussian primitives and ensure that the scales of these Gaussians are small enough. To validate these two conclusions, we designed an experiment to provide evidence from a one-dimensional rendering perspective.

\begin{figure}[htb]
\centering
 \includegraphics[width=0.95\linewidth]{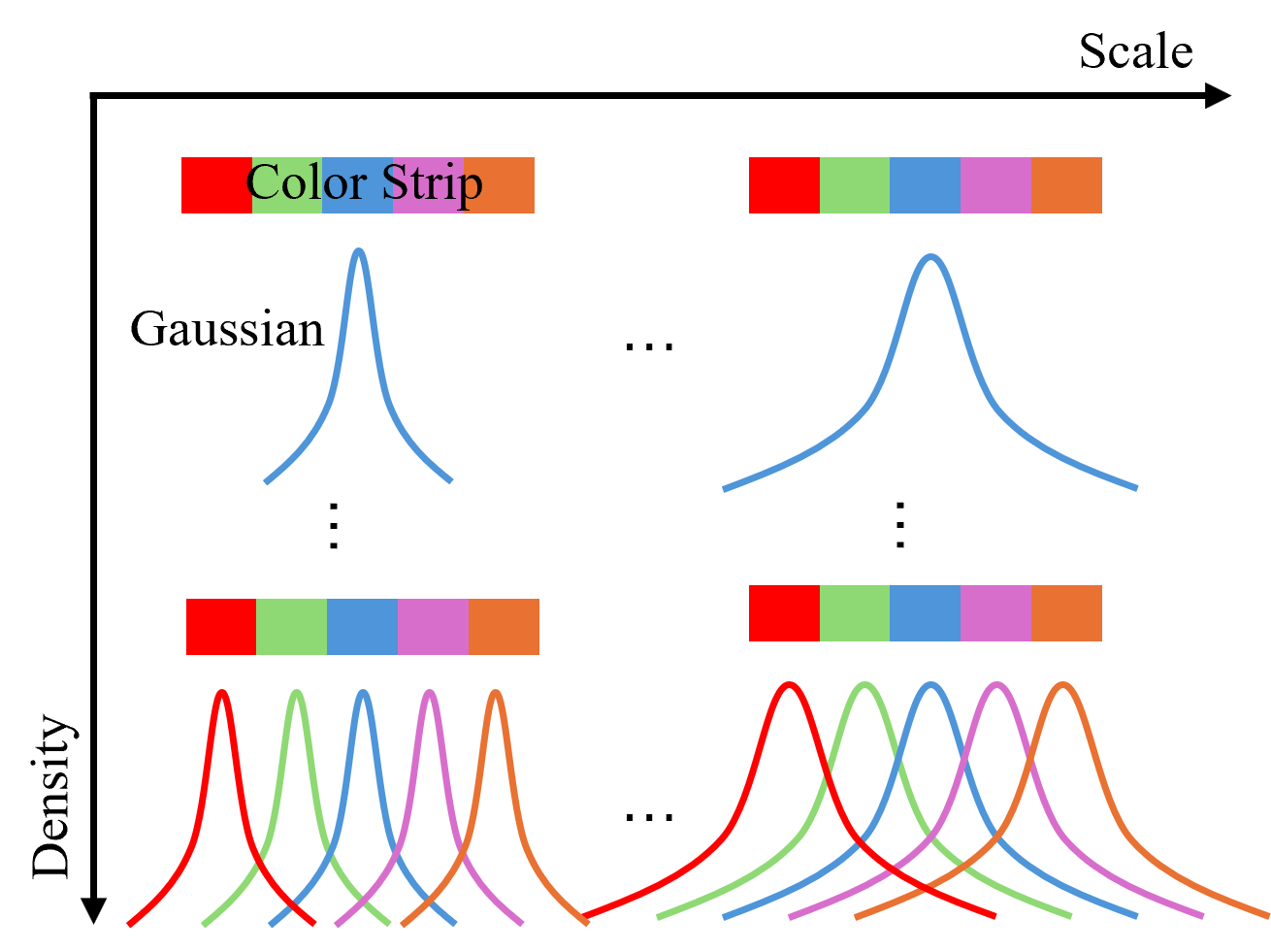}
 \caption{
   Illustration for the experiment of rendering the same color strip with different Gaussian arrangement.
 }
 \label{fig:signal_add_demo}
\end{figure}

\begin{figure*}[!htb]
 \includegraphics[width=0.9\linewidth]{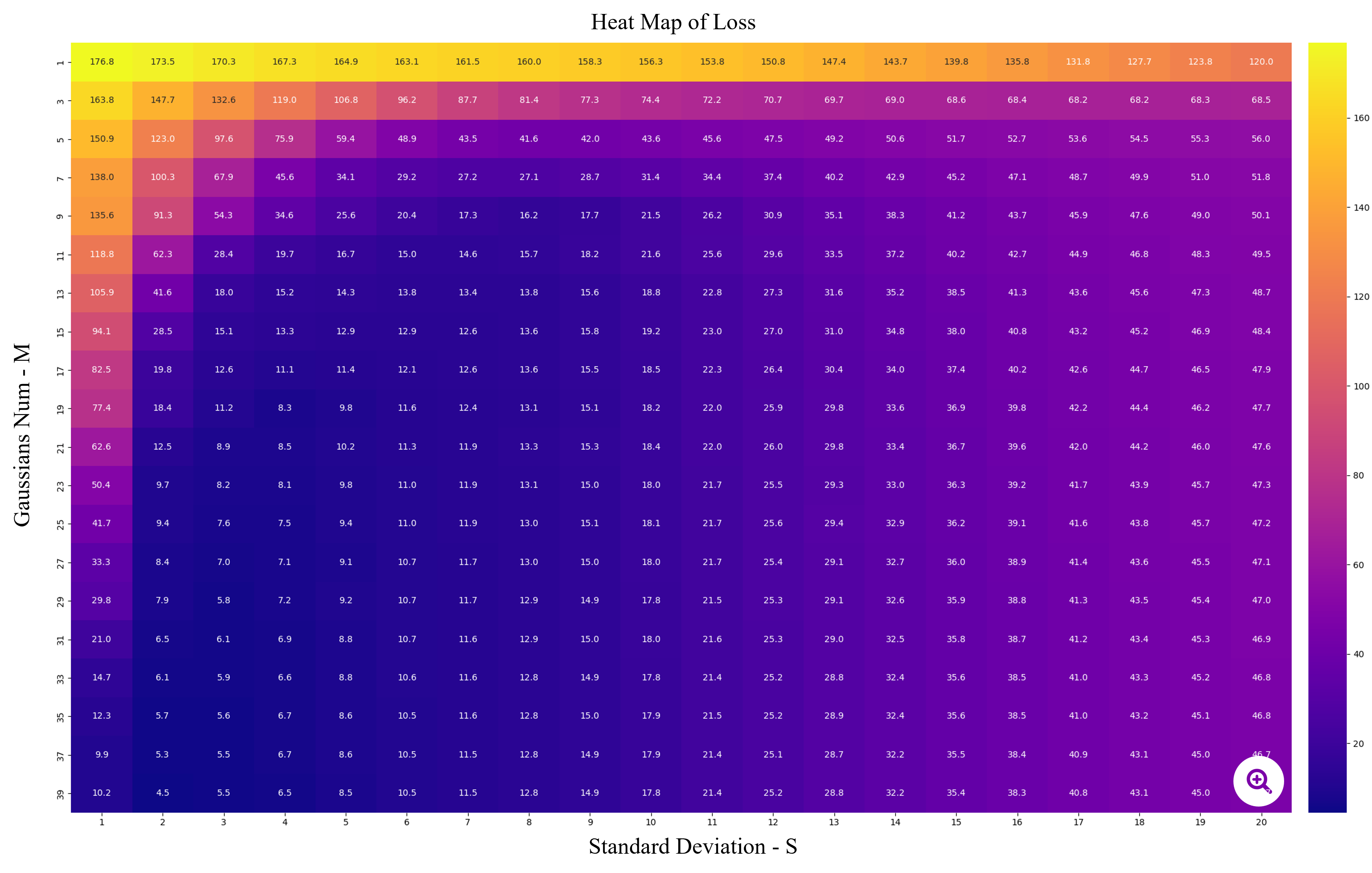}
 \caption{
   The color loss with different Gaussian numbers and standard deviations correspond to density and scale. The final loss is marked with white in every cell.
 }
 \label{fig:signal_add_result}
\end{figure*}

\textbf{Experiment Design:} We generated a striped color band composed of $ K $ different colors, with a shape of $[N, 3]$. Each color occupies $[N/K, 3]$ within the band. We then uniformly arranged $ M $ Gaussian models along this color band, where $ G_m \sim \mathcal{N}(f(M, m), S^2) $, with $ f(M, m) $ representing the mean of each Gaussian under a uniform distribution within the range $[0, N]$, and $ S $ being the standard deviation of each Gaussian. Each Gaussian is also associated with a learnable color parameter $\mathbf{c}$. As shown in Fig.~\ref{fig:signal_add_demo}. The colors of multiple Gaussians are superimposed to render the colors of the band, following the formula:
\begin{equation}
C_{render}(i) = \sum_{m=1}^{M}{\mathbf{c}_m G_m(i)},
\end{equation}
where $i$ is an integer position on the color strip $i \in [0, 100]$. The loss function is defined as:
\begin{equation}
\mathcal{L} = \frac{1}{N} \sum_{i=0}^{N}{| C_{gt}(i) - C_{render}(i) |}.
\end{equation}
We then varied the values of $M$ and $S$ within a certain range, recording the final color loss between the rendering color and the ground truth after the training had converged. The results of color loss are presented in the heatmap as illustrated in Fig.~\ref{fig:signal_add_result}.

\textbf{Experiment Analysis:} As shown in Fig. 2, when the number of Gaussians is fixed, for example, $ M = 15 $, as the standard deviation $ S $ gradually increases from 1 to 20, the loss decreases initially and then increases. The initial decrease occurs because when $ S $ is small, the current number of Gaussians is insufficient to represent the different colors in the color band. The loss reaches its lowest point at $ S = 7 $ with a value of 12.6, then increases again, indicating that larger scales are less effective at capturing high-frequency details than smaller scales. 

On the other hand, when the standard deviation of the Gaussians is fixed, for example, $ S = 6 $, the loss decreases as the number of Gaussians increases. However, if $ S $ remains unchanged, the loss eventually converges to a certain value, and further increasing the number of Gaussians does not lead to a significant decrease in loss but will increase the rendering burden, which means redundancy. At this point, the loss can be lowered only by further reducing the scale. For instance, when $ S = 6 $ and $ M = 37 $, lowering $ S $ to 2 results in a lower loss.
Note that, even with proper scale, the loss still will reach its optimum at a certain number of Gaussians, and any other increase of Gaussians only introduces redundancy.

From these observations, we can infer that to reduce the final rendering loss and better express a region with certain frequency information, we need to both increase the number of Gaussians and choose an appropriate corresponding scale. This indicates a balance between Density and Scale. Our method aims to establish a relation between them to maintain this balance and improve rendering performance.

\section{Details of Confidence Calculation}
Due to space limitations in the paper, we will provide additional details on the calculation of confidence here. First, we will reiterate some of the concepts mentioned in the paper.

We select several observable views for each Gaussian. For a given view $v_i$, we obtain the contribution $c$ of the Gaussian $G$ under this view by
\begin{equation}
    c(G, v_i) = \sum_{\mathbf{x}' \in \mathcal{F}_G^{v_i}}{\alpha G'(\mathbf{x}')},
\end{equation}
where $\mathcal{F}_G^{v_i}$ is the footprint of $G$ at view $v_i$ and $G'$ denotes its 2D Gaussian distribution. Then, we employ a CUDA-based Heap Sort to retrieve the top $M$ views with the largest contributions for the Gaussian $G$ as the observable views $V^{(M)}$:
\begin{equation}
    V^{(M)} = HeapSort_M({c(G, v_i)\,|\,i=1,2,...,N_v}),
\end{equation}
where $N_v$ is the total number of views.

Subsequently, we initialize the 8 directions by:
\begin{equation}
    \mathcal{D} = \{[\pm \frac{\sqrt{2}}{2}, \pm \frac{\sqrt{2}}{2}], [\pm 1, 0], [0, \pm 1]\}.
\end{equation}
In the following, we represent $\mathcal{D}$ as a matrix with shape $[8, 2]$. Then for the 2D Gaussian distribution $\Sigma' \in \mathcal{R}^{2 \times 2}$ obtained from each projection, given a sample radius $r$, we can compute the sample points as:
\begin{equation}
    \mathcal{S} = r\mathcal{D} \Sigma' + \mu'.
\end{equation}
The $\mu'$ is the projection center of Gaussian. The radius $r$ is picked from $\{0.5, 1.0, 1.5, 2.0, 2.5, 3.0\}$. Then along with the projection center $\mu'$, we get $49$ sample points, denoted as $\mathcal{S}_G^{i}$, where $i$ indicate the $i$-th view. Besides, we compute the Gaussian value for each sample point as the points' weight. Denoted as
\begin{equation}
    \Omega_G^{i} = G'(\mathcal{S}_G^{i}),
\end{equation}

After obtaining the sampling points and their weights for the Gaussian $G$ across all of the observable views $V^{(M)}$, we calculate the weighted SSIM between these sampling points:
\begin{equation}
    confidence = \frac{1}{M-1} \sum_{i=2}^{M}{SSIM(\mathcal{S}_G^{1}, \Omega_G^{1},\mathcal{S}_G^{i}, \Omega_G^{i})}.
\end{equation}
where ${S}_G^{1}$ corresponds to the view with the largest contribution and is regarded as a reference view. The weighted SSIM is computed as:
\begin{equation}\small
SSIM(\mathcal{S}_G^{1}, \Omega_G^{1},\mathcal{S}_G^{i}, \Omega_G^{i}) = \frac{(2\mu_1\mu_i + C_1) (2\sigma_{1i} + C_2)}{(\mu_1^2 + \mu_i^2 + C_1) (\sigma_1^2 + \sigma_i^2 + C_2)}
\end{equation}

\begin{equation}
\mu_1 = \frac{\sum_{p \in \mathcal{S}_G^{1}, \omega \in \Omega_G^{1}}\omega \cdot I_1(p)}{\sum_{\omega \in \Omega_G^{1}}\omega},
\end{equation}
\begin{equation}
\mu_i = \frac{\sum_{p \in \mathcal{S}_G^{i}, \omega \in \Omega_G^{i}}\omega \cdot I_i(p)}{\sum_{\omega \in \Omega_G^{i}}\omega},
\end{equation}
\begin{equation}
\sigma_1^2 = \frac{\sum_{p \in \mathcal{S}_G^{1}, \omega \in \Omega_G^{1}}\omega \cdot I_1(p)^2}{\sum_{\omega \in \Omega_G^{1}}\omega} - \mu_1^2, 
\end{equation}
\begin{equation}
\sigma_i^2 = \frac{\sum_{p \in \mathcal{S}_G^{i}, \omega \in \Omega_G^{i}}\omega \cdot I_i(p)^2}{\sum_{\omega \in \Omega_G^{i}}\omega} - \mu_i^2,
\end{equation}
\begin{equation}
\sigma_{1i} = \frac{\sum_{p_1 \in \mathcal{S}_G^{1}, p_i \in \mathcal{S}_G^{i}, \omega \in \Omega_G^{1}}\omega \cdot I_1(p_1) \cdot I_i(p_i)}{\sum_{\omega \in \Omega_G^{1}}\omega} - \mu_1 \mu_i,
\end{equation}
where $I_1(\cdot)$ and $I_i(\cdot)$ are functions to compute the color value ([0-255]) on the image according to the position. $C_1$ and $C_2$ are constant value, defined as $6.5025$ and $58.5225$ respectively.

\section{Quantitative Results}
This section will present the detailed data for each scenario in the experiment; see Tables 1-8.
Here, we only present the results of the methods we reproduced. Since FreGS is neither open-sourced nor provides detailed data for each scenario, we do not include a comparative display. Additionally, it is important to mention that the point cloud file ultimately saved by the Scaffold-GS method actually corresponds to the anchor points. The actual number of Gaussians involved in the rendering is multiplied by 5, which corresponds to the number of generated offsets. For more details, please refer to their paper. Therefore, the number of points listed in our paper for the Scaffold-GS method is multiplied by 5.

\begin{table*}[!htb]
\caption{SSIM scores for Mip-NeRF360 scenes. The higher, the better.}
\centering
\begin{tabular}{c|ccccc|cccc}
\hline
             & bicycle & flowers & garden & stump & treehill & room  & counter & kitchen & bonsai \\ \hline
3DGS         & 0.747   & 0.589   & 0.856  & 0.770 & 0.636    & 0.927 & 0.915   & 0.932   & 0.947  \\
Scaffold-GS  & 0.737   & 0.567   & 0.847  & 0.753 & 0.642    & 0.928 & 0.900   & 0.930   & 0.947  \\
MipSplatting & 0.780   & 0.627   & 0.868  & 0.791 & 0.639    & 0.933 & 0.920   & 0.935   & 0.951  \\
3D-HGS       & 0.752   & 0.587   & 0.859  & 0.761 & 0.633    & 0.929 & 0.919   & 0.935   & 0.945  \\ \hline
Ours         & 0.787   & 0.643   & 0.867  & 0.815 & 0.656    & 0.935 & 0.923   & 0.937   & 0.952  \\ \hline
\end{tabular}
\end{table*}

\begin{table*}[!htb]
\caption{PSNR scores for Mip-NeRF360 scenes. The higher, the better.}
\centering
\begin{tabular}{c|ccccc|cccc}
\hline
             & bicycle & flowers & garden & stump  & treehill & room   & counter & kitchen & bonsai \\ \hline
3DGS         & 25.124  & 21.402  & 27.265 & 26.681 & 22.547   & 31.536 & 29.105  & 31.409  & 32.291 \\
Scaffold-GS  & 25.058  & 21.161  & 27.322 & 26.435 & 23.201   & 32.133 & 28.701  & 31.527  & 32.601 \\
MipSplatting & 25.446  & 21.666  & 27.560 & 27.011 & 22.318   & 31.884 & 29.308  & 31.903  & 32.447 \\
3D-HGS       & 25.329  & 21.407  & 27.647 & 26.445 & 22.532   & 32.291 & 29.730  & 32.215  & 32.952 \\ \hline
Ours         & 25.600  & 21.790  & 27.754 & 27.616 & 22.681   & 32.174 & 29.293  & 32.016  & 32.604 \\ \hline
\end{tabular}
\end{table*}

\begin{table*}[!htb]
\caption{LPIPS scores for Mip-NeRF360 scenes. The lower, the better.}
\centering
\begin{tabular}{c|ccccc|cccc}
\hline
             & bicycle & flowers & garden & stump & treehill & room  & counter & kitchen & bonsai \\ \hline
3DGS         & 0.244   & 0.359   & 0.122  & 0.242 & 0.347    & 0.197 & 0.184   & 0.116   & 0.180  \\
Scaffold-GS  & 0.266   & 0.383   & 0.142  & 0.276 & 0.352    & 0.197 & 0.213   & 0.121   & 0.185  \\
MipSplatting & 0.188   & 0.291   & 0.106  & 0.206 & 0.295    & 0.180 & 0.170   & 0.110   & 0.165  \\
3D-HGS       & 0.234   & 0.360   & 0.119  & 0.243 & 0.345    & 0.193 & 0.178   & 0.113   & 0.177  \\ \hline
Ours         & 0.189   & 0.259   & 0.112  & 0.190 & 0.267    & 0.176 & 0.161   & 0.107   & 0.160  \\ \hline
\end{tabular}
\end{table*}

\begin{table*}[!htb]
\caption{The number of Gaussians for Mip-NeRF360 scenes. The lower, the better.}
\centering
\begin{tabular}{c|ccccc|cccc}
\hline
             & bicycle & flowers & garden  & stump   & treehill & room    & counter & kitchen & bonsai  \\ \hline
3DGS         & 5764211 & 3425173 & 5688534 & 4567380 & 3438184  & 1509242 & 1171832 & 1760622 & 1245115 \\
Scaffold-GS  & 5031965 & 4197115 & 4743785 & 4042250 & 4260260  & 1674495 & 1670550 & 1804580 & 2716075 \\
MipSplatting & 8394174 & 4701228 & 6761105 & 5839380 & 4994575  & 2117080 & 1484976 & 2141862 & 1621641 \\
3D-HGS       & 5361002 & 3199773 & 4878283 & 4095968 & 3520583  & 1348901 & 1083763 & 1625120 & 1106191 \\ \hline
Ours         & 3164257 & 4134060 & 2378607 & 3200223 & 3818096  & 1277790 & 1166784 & 1138818 & 1050370 \\ \hline
\end{tabular}
\end{table*}

\begin{table*}[!htb]
\caption{SSIM scores for Tanks \& Temples and Deep Blending scenes. The higher, the better.}
\centering
\begin{tabular}{c|cc|cc}
\hline
             & train & truck & drjohnson & playroom \\ \hline
3DGS         & 0.818 & 0.878 & 0.898     & 0.901    \\
Scaffold-GS  & 0.816 & 0.883 & 0.907     & 0.910    \\
MipSplatting & 0.830 & 0.888 & 0.899     & 0.907    \\
3D-HGS       & 0.831 & 0.886 & 0.902     & 0.905    \\ \hline
Ours         & 0.837 & 0.889 & 0.905     & 0.906    \\ \hline
\end{tabular}
\end{table*}

\begin{table*}[!htb]
\caption{PSNR scores for Tanks \& Temples and Deep Blending scenes. The higher, the better.}
\centering
\begin{tabular}{c|cc|cc}
\hline
             & train  & truck  & drjohnson & playroom \\ \hline
3DGS         & 22.237 & 25.536 & 29.046    & 29.916   \\
Scaffold-GS  & 22.470 & 25.932 & 29.744    & 30.654   \\
MipSplatting & 22.096 & 25.760 & 28.820    & 29.866   \\
3D-HGS       & 23.037 & 26.261 & 29.162    & 29.709   \\ \hline
Ours         & 22.422 & 26.110 & 29.529    & 30.434   \\ \hline
\end{tabular}
\end{table*}

\begin{table*}[!htb]
\caption{LPIPS scores for Tanks \& Temples and Deep Blending scenes. The lower, the better.}
\centering
\begin{tabular}{c|cc|cc}
\hline
             & train & truck & drjohnson & playroom \\ \hline
3DGS         & 0.206 & 0.155 & 0.246     & 0.247    \\
Scaffold-GS  & 0.219 & 0.152 & 0.253     & 0.255    \\
MipSplatting & 0.189 & 0.133 & 0.242     & 0.236    \\
3D-HGS       & 0.198 & 0.148 & 0.240     & 0.248    \\ \hline
Ours         & 0.161 & 0.120 & 0.228     & 0.234    \\ \hline
\end{tabular}
\end{table*}

\begin{table*}[!htb]
\caption{The number of Gaussians for Tanks \& Temples and Deep Blending scenes. The lower, the better.}
\centering
\begin{tabular}{c|cc|cc}
\hline
             & train   & truck   & drjohnson & playroom \\ \hline
3DGS         & 829641  & 1733197 & 3272047   & 2333595  \\
Scaffold-GS  & 831215  & 1313735 & 1128830   & 998420   \\
MipSplatting & 1148800 & 2333585 & 4156156   & 2829034  \\
3D-HGS       & 825246  & 1613097 & 3255567   & 2015244  \\ \hline
Ours         & 2604486 & 1531760 & 2459542   & 919977   \\ \hline
\end{tabular}
\end{table*}

\section{Iteration Details}
\begin{figure*}[htb]
 \includegraphics[width=\linewidth]{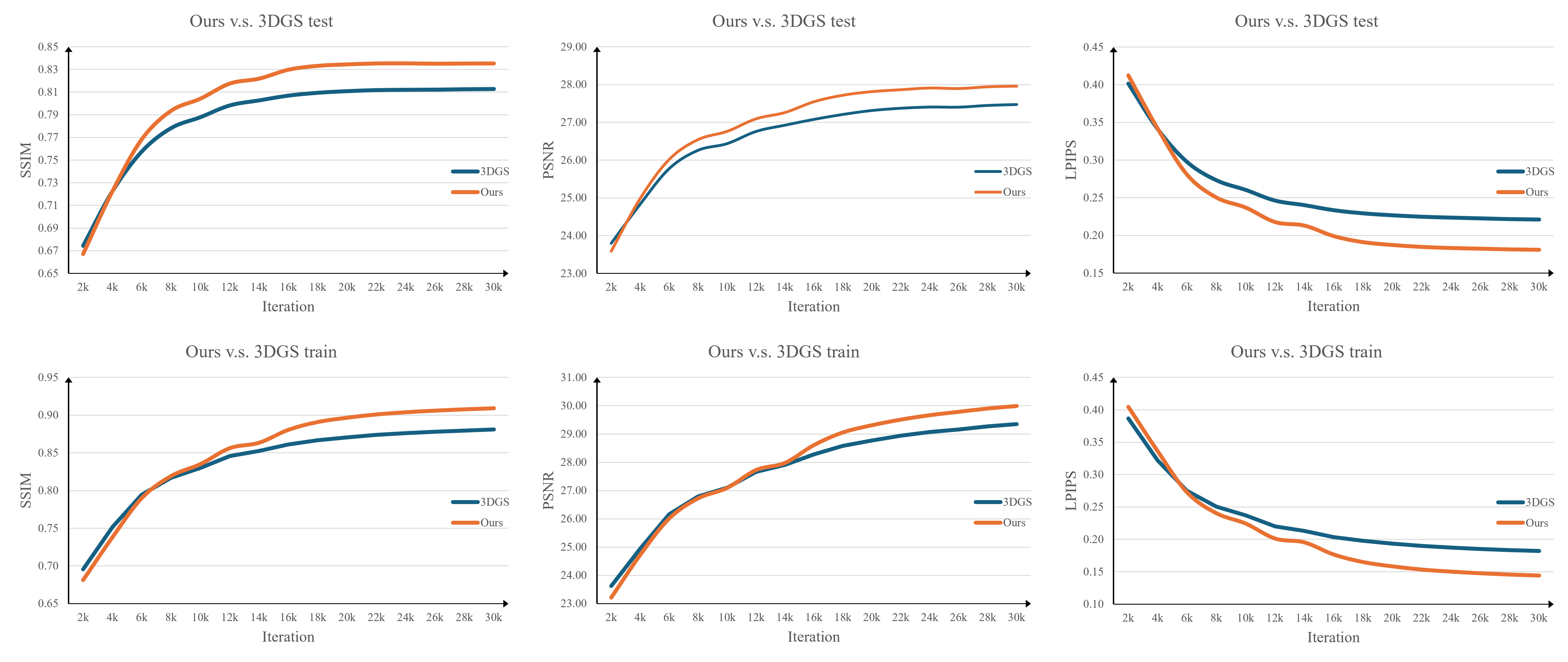}
 \caption{
   Convergence curves of our FDS-GS method and 3DGS over 30,000 iterations.
 }
 \label{fig:converge_speed}
\end{figure*}

In this section, we will provide the convergence curves for our FDS-GS method and 3DGS during the 30,000 iteration process, showing the results every 2,000 steps for both the training and test sets. Our experiment is tested on \textit{MipNeRF360} dataset with all scenes. As shown in Fig.~\ref{fig:converge_speed}, our FDS-GS method not only has better convergence on the training set but also performs more robustly on the test set, which fully demonstrates the robustness of our method.

\section{More Qualitative Results}
\begin{figure*}[htb]
 \includegraphics[width=\linewidth]{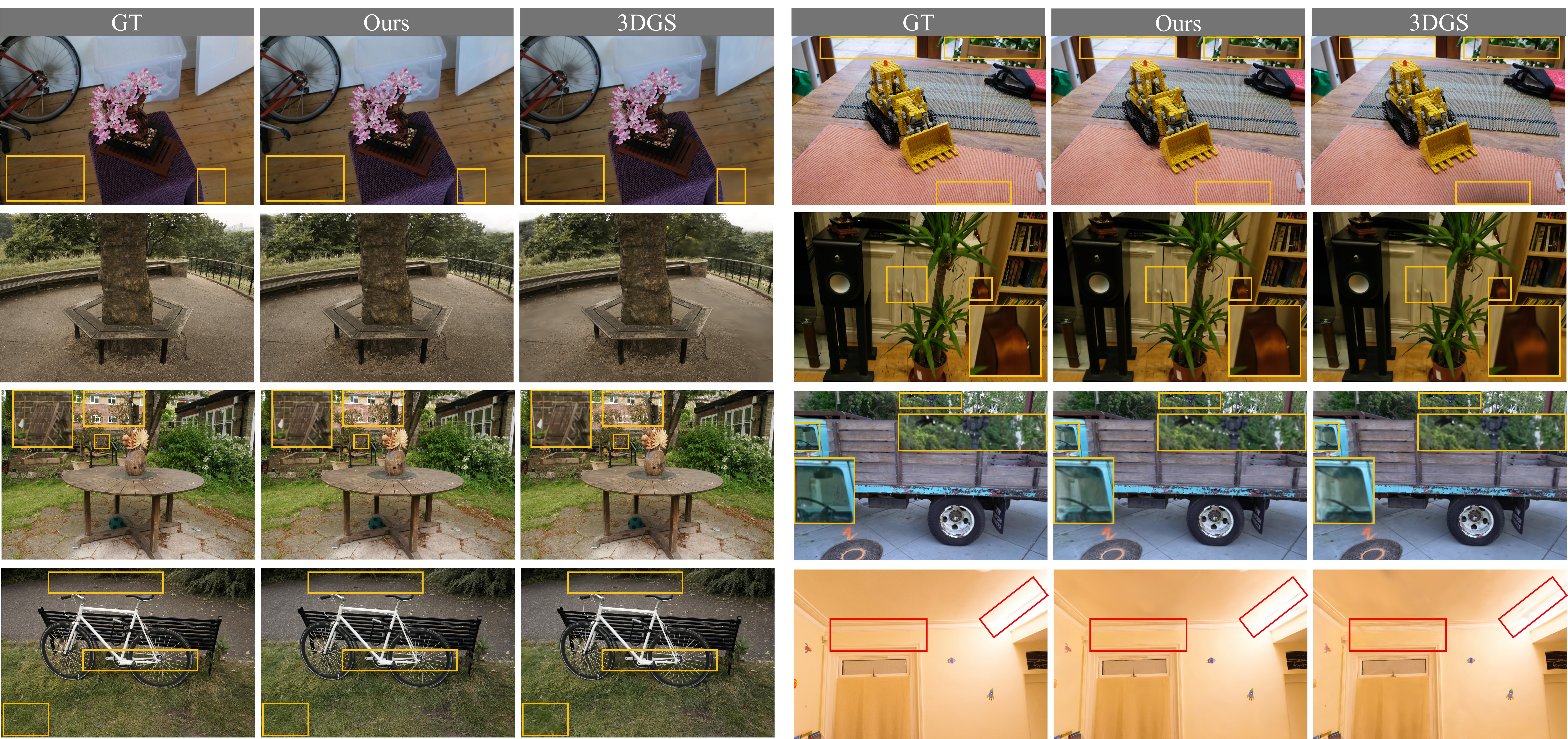}
 \caption{
   Qualitative comparison of our FDS-GS method with the vanilla 3DGS. Our method provides a superior restoration of scene details. Zoom in for more details.
 }
 \label{fig:experiment_vis_supp}
\end{figure*}

In this section, we provide more qualitative comparison results, see Fig.~\ref{fig:experiment_vis_supp}. Our FDS-GS method provides a superior restoration of scene details than the vanilla 3DGS. 

